\documentclass[a4paper, 10pt, conference]{ieeeconf}      

\IEEEoverridecommandlockouts                              

\usepackage{graphics} 
\usepackage{epsfig} 
\usepackage{mathptmx} 
\usepackage{times} 
\usepackage{amsmath} 
\usepackage{amssymb}  
\usepackage{bm}  

\usepackage{color,soul}

\title{\LARGE \bf
Kinematics Transformer: Solving The Inverse Modeling Problem of Soft Robots using Transformers
}

\author{Abdelrahman Alkhodary$^{1}$ and Berke Gur$^{2}$
\thanks{$^{1}$Abdelrahman Alkhodary is with the Graduate School of Engineering, Department of Computer Engineering, Bahcesehir University, Istanbul, Turkey
        {\tt\small abdelrahman.alkhoda@bahcesehir.edu.tr}}%
\thanks{$^{2}$Berke Gur is with Faculty of  Engineering and Natural Sciences, Department of Mechatronics Engineering, Bahcesehir University, Istanbul, Turkey
        {\tt\small berke.gur@eng.bau.edu.tr}}%
}

\begin{document}

\maketitle
\thispagestyle{empty}
\pagestyle{empty}

\begin{abstract}

Soft robotic manipulators provide numerous advantages over conventional rigid manipulators in fragile environments such as the marine environment. However, developing analytic inverse models necessary for shape, motion, and force control of such robots remains a challenging problem. As an alternative to analytic models, numerical models can be learned using powerful machine learned methods. In this paper, the Kinematics Transformer is proposed for developing accurate and precise inverse kinematic models of soft robotic limbs. The proposed method re-casts the inverse kinematics problem as a sequential prediction problem and is based on the transformer architecture. Numerical simulations reveal that the proposed method can effectively be used in controlling a soft limb. Benchmark studies also reveal that the proposed method has better accuracy and precision compared to the baseline feed-forward neural network.     
\end{abstract}

\section{INTRODUCTION}\label{sec: Intro}

Fueled by the rapid population increase and the associated over-consumption of the 20th century, many experts believe that the world is heading for a resource crisis. It is predicted that the existing natural resources will be insufficient to meet our daily needs in critical areas such as food and energy in the very near future. Higher demand on natural resources is forcing policy makers to consider increasing resource utilization, reducing waste, and improving recycling efficiency. In addition, the sustainable exploitation of the vast but mostly untouched oceans and seas is also another option. The marine environment is becoming a critical source of food, proteins, hydrocarbons and fossil fuels, alternative energy sources, mines and minerals. In this respect, item 14 of the United Nations sustainable development goals addresses this issue, and aims to promote the conservation and sustainably utilization of the oceans, seas and marine resources for sustainable development \cite{UN15}. 

However, the underwater environment is not a suitable for humans to live or engage in any prolonged activity. The water pressure increases by 1 bar approximately every 10m, having adverse affects on human physiology. Our vision and aural abilities are also vastly degraded in the underwater environment, making it difficult to see and to communicate. Hydrodynamics of the underwater environment is very different than and unpredictable compared to  terrestrial dynamics, causing humans to loose their dexterity. These problems make the utilization of unmanned underwater robots very attractive and often necessary for sub-sea exploration and intervention activities.  

Since their first deployment in the 1970’s, unmanned underwater vehicles have evolved very little. Current worker class remotely operated vehicles (ROV) can host one or more heavy-duty, hydraulically actuated manipulators for underwater manipulation. Despite being powerful, such manipulators are generally clumsy and lack the precision and dexterity required for operating in delicate and fragile marine environments. To overcome this shortcoming, electro-mechanically actuated underwater manipulators are developed and deployed \cite{Kha16}. 

As an alternative to rigid electro-mechanical underwater manipulators hyper-elastic soft manipulators have gained significant research interest in the past decade. Compared to their electro-mechanical counterparts, soft manipulators generally provide low performance at a low cost. However, soft manipulators make up for this low performance through their compliance, eliminating the need for high-precision motion and force control strategies, while making them very suitable for operating in fragile environments. Unfortunately, due to the continuum nature of these soft manipulators, developing accurate forward analytical models that map the actuator space to the task space is challenging, making inverse analytical modeling intractable. Furthermore, established methods for modeling conventional rigid manipulators are not applicable to soft and deformable robots. Therefore, developing effective and efficient inverse models for soft manipulators remains an elusive problem.       

With this paper, a transformer architecture based approach for machine learning the inverse kinematics of soft robotic limbs is presented. In line with robotics jargon, the term `kinematics' here refers to the steady-state configuration of the limb. To the best knowledge of the authors, this paper represented the first attempt to use the transformer architecture-based machine learning methods for shape and motion control of soft robots. The remainder of the paper is organized as follows. A concise review of existing literature on machine learning of soft robotic arm models is presented in Sec. \ref{sec: LitRev}. This is followed by Sec. \ref{sec: LimbModel} where the soft robotic limb model used in this research is introduced. Next, the proposed inverse model based on the transformer architecture, namely, the Kinematics Transformer, is described in Sec. \ref{sec: InvModel}. Results pertaining to the numerical evaluation of the Kinematics Transformer is presented in Sec. \ref{sec: Results}. The paper is concluded with a brief discussion of the results and future work in Sec. \ref{sec: Conc}. 

\section{RELATED WORK}\label{sec: LitRev}

Holsten et. al. \cite{Hol19} learned the inverse kinematics model of a soft robot arm using polynomial regression, where the input is shape of the arm and the output is the cable displacement that control the soft arm. The shape of the arm is represented as points distributed over the soft arm.
While the authors in \cite{Ber20} first learned the forward kinematics model from the obtained data of the arm, then used that model to learn the inverse model by leveraging an optimization algorithm. The gradient-based optimization is used to search the learned forward kinematics neural network by minimizing the distance between the desired position for the end-effector and the output of the inverse kinematics neural network of which corresponds to the control inputs.
Morgan et. al. \cite{Gil18} trained a neural network to learn the nonlinear dynamics of the soft arm. Then used the gradient of the neural network with respect to each input to construct the state matrix, and then used the gradient of the network with respect to only the control signals to construct the input matrix. They used the state matrix and input matrix to apply model predictive control to control the arm.
The authors of \cite{Kov19} investigated the performance of applying the inverse kinematics through neural network for a soft arm. The soft arm of building blocks that each has three identical hollow cylinders, passing through them three rubber hoses. The arm is pneumatically actuated with hydraulic pressure. The inverse kinematics model learns the mapping between the amount of pressure applied to each rubber hose and the position of the tip of the arm. In \cite{Thu16} Gillespie  et. al. attempt to learn the inverse kinematics model of a continuum robot. The data used to train the inverse neural network model was generated from a kinematic model that uses a constant curvature approximation for modeling the continuum kinematics of the arm. The arm consists three segments with each segment controlled by three pneumatic actuators. The input to the inverse model is the current actuations of the segments along with the desired position of the tip of the arm, while the output would be the actuations of the segment to reach the desired position of the arm. 

The feed-forward neural networks (FFNN) are designed to learn to map inputs to outputs. Therefore, such networks are very suitable for learning tasks such as regression. For example, in learning the inverse kinematics of robotic limbs FFNN's are used to map the desired tip position to the corresponding actuation inputs. However, FFNN's cannot keep track of sequential data, and thus are not very suitable for learning sequential problems such the time-dependent inverse dynamic modeling of the limb motion.

The inverse kinematics problem can be cast into a sequential decision making problem by segmenting the path between the current state (e.g., tip position) and the final desired state into intermediate way-points. As stated above, the inverse dynamics problem involves modeling time-evolution of the trajectory of the limb and therefore, by definition is sequential. For such sequential problems, alternative architectures and algorithms that stem from other areas of machine learning such as natural language processing are more suitable. This paper proposes an architecture called Kinematics Transformer (KT) to solve the inverse kinematics problem based on the transformer architecture. The KT model can be used in solving the inverse kinematics problem by reaching a single desired position or multiple way points as it have the advantages of considering the past positions it has reached.

The transformer architecture was first introduced by Vaswani et. al. \cite{Vas17} for language translation. The core of the transformer architecture is the attention mechanism. The transformer architecture has an encoder and decoder parts, while the encoder has self-attention mechanism and the decoder has Mask self-attention mechanism. Since the introduction of the transformer architecture most of the recent state of the art models in natural language processing is based on it. One of the use cases BERT \cite{Dev19} architecture is based on the encoder part of the transformer and is used for question answering where a sentence is fed to the architecture with a Mask token for the missing word to the architecture to find. As the name of the GPT \cite{Rad18} Generative Pre-trained Transformer, the GPT architectures are based on the decoder part of the transformer and used to generate a token based on the input sequence. Transformer architecture has been also proposed in \cite{Dos21} for image classification and named vision transformer. As transformer can deal with sequence not an image, the authors of vision transformer proposed to crop the input image into a set of patches, then each patch will be flattened and linearly projected into a fixed dimension. Chen et. al. \cite{Che19} proposed a new framework, namely the decision transformer, that casts reinforcement learning as a conditional sequence modeling problem and utilizes the transformer architecture.

\section{SOFT ROBOTIC LIMBS}\label{sec: LimbModel}


Inspired by the arms and tentacles of squid, a slender solid frustum shaped geometry is assumed for the limbs. The Cosserat theory is adapted for developing analytical models of the limbs due to its versatility. In this section, details regarding the analytic limb model used in this study is provided.

\subsection{Analytical Limb Model}
\begin{figure}[htbp]
\centerline{\includegraphics[width=80mm]{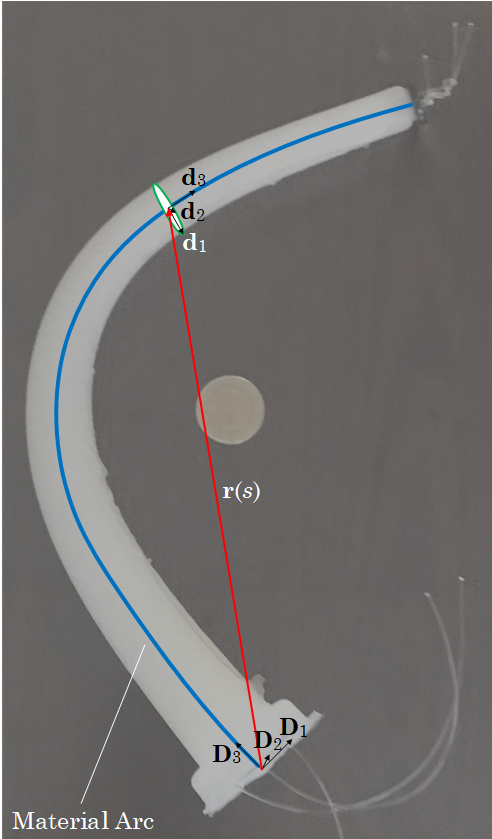}}
\caption{A picture of an early prototype half-scale soft robotic limb depicting the material arc, the inertial and local coordinate systems, as well as the task space position vector $\mathbf{r}(s)$.}
\label{fig: halfScaleLimb}
\end{figure}

Following the approach presented by Ren et. al., an analytic kinematic model for the limbs is developed based on the Cosserat model \cite{Ren12}. Accordingly, the shape of a limb of nominal length $L$ is described using the material arc length coordinate $0 \leq s \leq L$ that passes through the centroids of the circular cross sections along the longitudinal axis of the limb. In addition to an inertial coordinate system fixed to the base of the limb (with unit vectors $\mathbf{D}_1$, $\mathbf{D}_2$, and $\mathbf{D}_3$) local coordinate systems attached to the cross-sections are used to describe the state of arm in the task-space in terms of position $\mathbf{r}(s)$ and orientation $\mathbf{R}(s)$. These local coordinate systems are defined by the two principle axes (with corresponding unit vectors $\mathbf{d}_1$ and $\mathbf{d}_2$) as well as an axis perpendicular to the cross-section (with a corresponding unit vector $\mathbf{d}_3$). The basic geometric descriptors of the limb relevant for the Cosserat model is presented in Fig. \ref{fig: halfScaleLimb}.

The state of the limb in the configuration space is described in terms of the longitudinal strain $\epsilon_3(s)$ and three curvatures $\mathbf{\kappa}(s) = [\kappa_1(s)$ $\kappa_2(s)$ $\kappa_3(s)]^T$, defined about the corresponding local coordinate axes. In this model, the Bernoulli-Euler beam assumption is exploited, where the cross-sections are assumed to remain in-plane, making the shear strains $\epsilon_1$ and $\epsilon_2$ equal to zero. The relationship between the longitudinal strain and curvatures is summarized in the expression

$$
\mathbf{d}_i^\prime (s) = [1+\epsilon_3(s)] \cdot \mathbf{\kappa}(s) \times \mathbf{d}_i(s) 
\eqno{(1)}
$$

\noindent for $i = 1,2,3$ and where $\cdot^\prime$ represents the spatial derivative with respect to $s$. 

The steady-state governing equations of the configuration are obtained from Newtonian mechanics as

$$
\mathbf{n}^\prime (s) + \mathbf{f}(s) = \mathbf{0}
\eqno{(2)}
$$

\noindent and 

$$
\mathbf{m}^\prime (s) + \mathbf{l}(s) + \mathbf{r}^\prime(s) \times \mathbf{n}(s)= \mathbf{0}
\eqno{(3)}
$$

\noindent where $\mathbf{n}$ and $\mathbf{m}$ are the internal forces and moments, respectively. In (2) and (3), $\mathbf{f}$ and $\mathbf{l}$ represent the external forces and moments, respectively, acting on the limb per unit length. These external forces and moments include contributions from gravity, buoyancy, and tendon loads. All terms expressed in (2) and (3) are defined with respect to the inertial coordinate system attached to the base of the limb.

Assuming linear-elastic behavior, the constitutive relations for the Cosserat limb model are expressed as:

$$
\begin{bmatrix}
EI_1(s) & 0 & 0 \\
0 & EI_2(s) & 0 \\
0 & 0 & GI_3(s) \\
\end{bmatrix} \cdot \mathbf{\kappa}(s) = \mathbf{m}(s)
\eqno{(4)}
$$

\noindent and

$$
\rho A(s) \cdot \epsilon_3(s) = n_3(s)
\eqno{(5)}
$$

\noindent where $E$, $G$, $\rho$, and $A$ are the Young's modulus, the modulus of rigidity, the mass density, and the cross-sectional area of the limb, respectively. The terms $I_i$ represent the moment of inertia of the cross-section with respect to axis $\mathbf{d}_i$, and $n_3$ is the component of $\mathbf{n}$ perpendicular to the cross-section. 

Equations (1)-(5) constitute the kinematic Cosserat model of the soft limbs utilized in forward kinematics calculations and dataset generation. Explicit expressions of the external forces $\mathbf{f}$ and moments $\mathbf{l}$ can be obtained based on tendon geometry. A two-stage numerical scheme outlined in \cite{Ren12} is employed for computing the forward kinematics solution where the first stage involves solving for the linear strain $\epsilon_3$ and curvatures $\mathbf{\kappa}$ for a given set of tendon forces from tip to base using (2)-(5). In the second stage, the position $\mathbf{r}$ and orientation $\mathbf{R}$ are calculated for the previously computed configuration space parameters using (1).     

\subsection{Limb Model Parameters}
The slender solid frustum shaped limbs are assumed to have a nominal length $L = 600$ mm. At the base, the cross-sectional diameter of the limb is 60mm and linearly reduces to 20mm at the tip. Although the actual squid limbs have more complex geometries and include distal enlarged sections called clubs as well as suckers, a simplified geometry is preferred for shape control research. The limbs are assumed to be manufactured using silicon-rubber material with a mass density of 1070 kg/m$^3$ Therefore the limbs are almost neutrally buoyant in fresh or salt-water. In addition, the silicon rubber material is assumed to have a modulus of elasticity of 70 kPa, a tensile strength of 1.40 MPa, and a shore hardness of 00-30. A single rigid disc is embedded into the distal end of the limb. These disks are 3-D printed from ABS material. A set of four tendons are connected to this disc at the distal end of the limb at a radial distance of 3.25 mm from the centerline. The radial distance of the tendons reach 20mm at the based. The limbs are actuated by the tension applied to these tendons by the actuators located at the base of the limb.

\subsection{Dataset Generation}\label{ssec: dataset}

To train the KT, a forward kinematics dataset was generated from the analytical model of the soft limb described in the previous section. The input to this forward model is the forces in the tendons and the output is the coordinates of the discrete points of the material curve in 3-D space. Uniformly random distributed tension forces in the range of $0 \leq T_i \leq 10$ N is generated for each tendon $i$. The corresponding tip position of the limb computed from this forward kinematics model is recorded to form a labeled training dataset. Data is generated in form of episodes, where each episode consists of 200 steps. Each episode is initialized at the nominal state of the soft limb with all tendon forces equal to zero and the limb is in the undeformed configuration with the tip at the rest position. The statistics of the training dataset is presented in Table \ref{Tab: TrainDataSet}.

\begin{table}[h]
\caption{Statistics of the training dataset. All results are presented in mm.}
\label{Tab: TrainDataSet}
\begin{center}
\setlength{\tabcolsep}{3pt}
\begin{tabular}{|c|c|c|c|c|c|}
\hline
Metric & $x$-coord.  & $y$-coord. & $z$-coord. & dist. from base & dist. from rest \\ 
\hline
Min. & $95.1$ & $-306.2$ & $-305.2$ & $184.4$ & $0.0 $ \\ 
\hline
Max. & $610.0$ & $305.4$ & $305.5$ & $610.0$ & $545.4$ \\
\hline
Mean & $308.1$ & $0.0$ & $0.0$ & $391.2$ & $375.5$ \\
\hline
STD & $114.1$ & $159.6$ & $159.4$ & $75.9$ & $118.5$ \\
\hline
\end{tabular}
\end{center}
\end{table}

\section{LEARNING THE KINEMATICS OF THE LIMB}\label{sec: InvModel}

The primary contribution of this paper, namely the KT, is presented in this section.

\subsection{Inverse Kinematics as a Sequence}

\begin{figure}[htbp]
\centerline{\includegraphics[width=\linewidth]{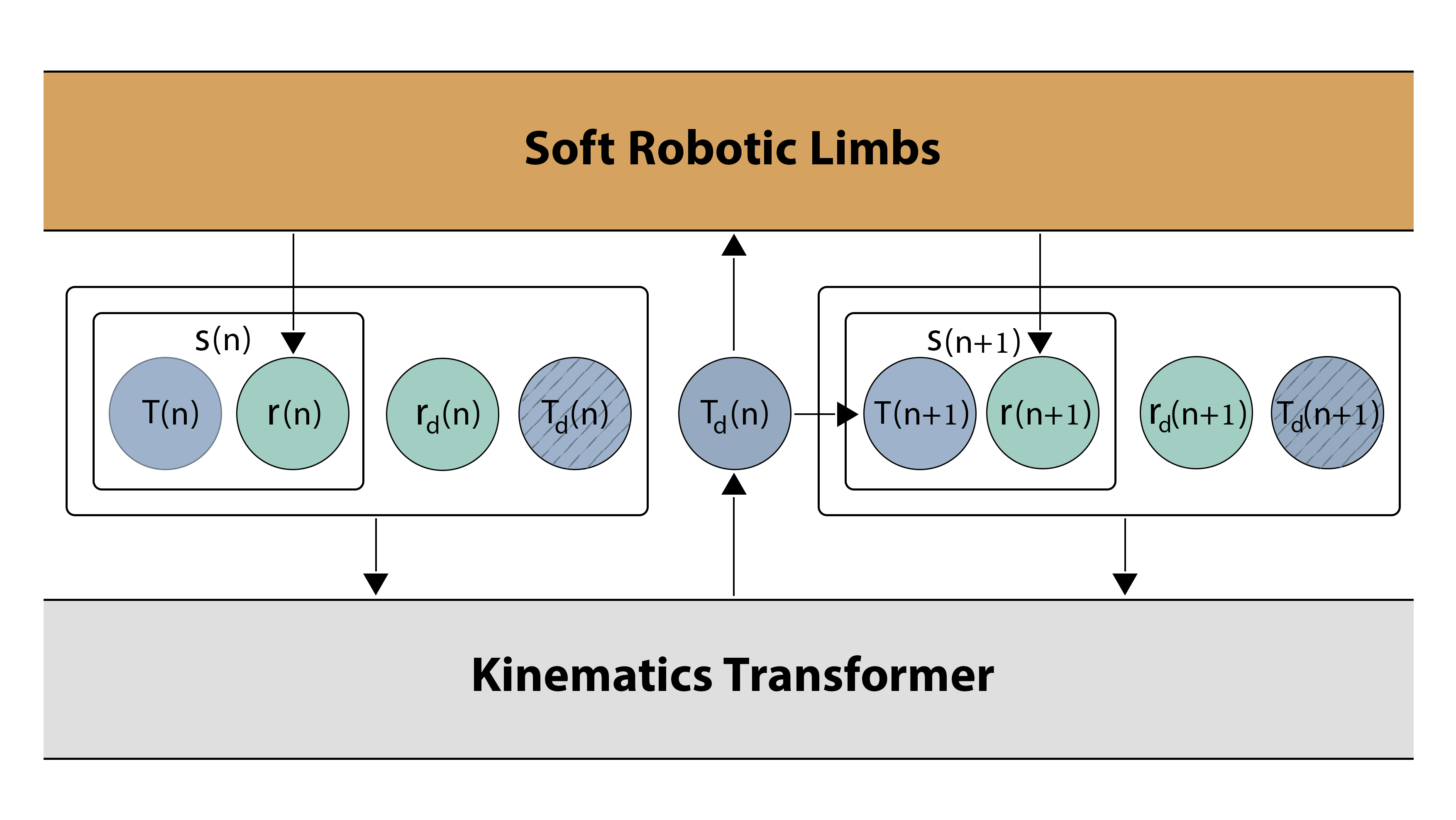}}
\caption{The token structure used in the KT with hashed tokens are masked tokens.}
\label{fig: sequence}
\end{figure}

The primary motivation for developing the KT was to effectively learn the inverse kinematics of soft robotic limbs in reaching a single point in 3-D space or reaching multiple points in a sequential order.  A sequence is represented as an $N$-element list of tokens $\bm{\Sigma} = [\bm{\tau}(0), \bm{\tau}(1), \cdots, \bm{\tau}(N-1)]$ where each token is defined as $\bm{\tau}(n) = [\mathbf{s}(n), \mathbf{r}_d(n), \mathbf{T}_d(n)]$ for $0 \leq n \leq N-1$. The current state of the limb is expressed as $\mathbf{s}(n) = [\mathbf{r}(n), \mathbf{T}(n)]$ with $\mathbf{r}(n)$ and $\mathbf{T}(n)$ representing the current tip position and current tendon forces, respectively. As shown in Fig.\ref{fig: sequence}, other elements of the token include the goal or desired tip position $\mathbf{r}_d(n)$ and the desired (but unknown) tendon forces $\mathbf{T}_d(n)$. The desired tendon forces $\mathbf{T}_d(n)$ are masked with zero, and are predicted by the model. These predicted tendon forces are applied to the limb and become the actual tendon forces in the next step, i.e., $\mathbf{T}(n+1) = \mathbf{T}_d(n)$. Note that the current desired tip position $\mathbf{r}_d(n)$ is generally different than the next actual tip position $\mathbf{r}(n+1)$ because the limb cannot reach the target tip position in one step.

\subsection{The Kinematic Transformer} \label{subsection-kt}

\begin{figure}[htbp]
\centerline{\includegraphics[width=80mm]{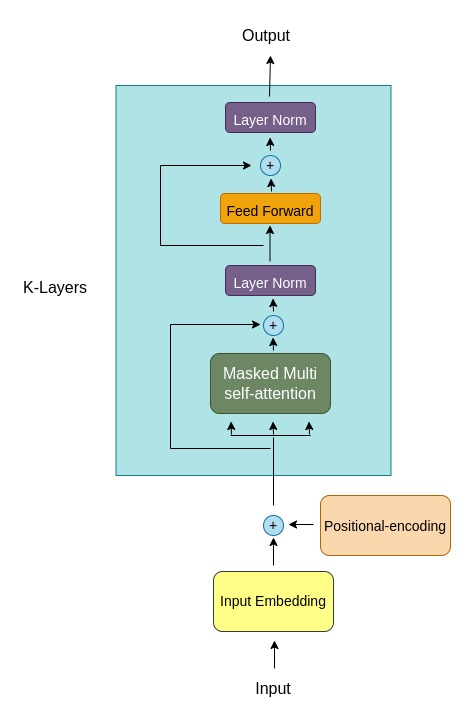}}
\caption{The GPT architecture used in the KT.}
\label{fig: gpt}
\end{figure}

Similar to the Decision Transformer, the base architecture of the KT is the Generative Pre-Trained Transformer (GPT) model. GPT is a $k$-layer decoder-only transformer. Each decoder layer is comprised of a masked multi-head self-attention mechanism as the first sub-layer and a position-wise fully connected second sub-layer, as shown in the Fig. \ref{fig: gpt} \cite{Rad18}. GPT models are auto-regressive models in the sense that they generate a new token based on past input tokens. The auto-regressive of the model comes from the masked self-attention layer. GPT architecture is designed to handle one token type and one token at each step. As is in the Decision Transformer, the input embeddings and the positional encodings are modified such that the embeddings of the three tokens $[\mathbf{s}(n), \mathbf{r}_d(n), \mathbf{T}(n)]$ are stacked. A linear layer is learned for obtain the embedding of each input token. Another linear layer is learned for the positional encodings at each step. The positional encoding layer output is added to the embedding of the three tokens.   


\subsection{Self Attention} \label{subsection-attention}

The masked self-attention mechanism is one of two sub-layers of the GPT transformer architecture.  The mask used to predict the token at step $n$ limits the GPT to only consider the tokens from the start of the sequence until step $n-1$, and not the entire sequence. The output of the masked self-attention sub-layer is a context richer substitution of the input sequence. The self-attention sub-layer is what gives the transformer the ability to effectively process long sequences. 

After passing through the embedding and positional encoding layers, the state $\mathbf{s}(n)$, the desired tip position $\mathbf{r}_d(n)$, and the tendon forces $\mathbf{T}_d(n)$ are stacked together and treated as one token. Each such token in the sequence will be fed to three one-layer neural networks that will generate three vectors, namely the Query $Q$, the Key $K$, and the Value $v$ vectors.  The dimension $d$ of these three vectors is the same as the dimension of the output of the embedding and positional encoding layers.  This  embedding (or hidden) dimension $d$ is a hyper-parameter that is determined by the user. The Query vectors are multiplied by the Key vectors to get a similarity score. A scaling of 
$\sqrt d$ is applied. To normalize the output a softmax layer is employed. In the final step, the output of softmax normalization is multiplied with the Value vectors. The computation of the attention is defined in (6).

$$
\text{Attention} = \text{softmax}(\mathbf{Q} \cdot \mathbf{K}^T / \sqrt{d}) \eqno{(6)}
$$

In general, the complete Query, Key, and Value vectors are not passed through the attention layer all at once. Instead, these vectors are split into $M$ groups, and each group is fed into a separate instance of the attention layer, running in parallel. These attention layers are called attention heads. Attention results from each attention head are concatenated together to produce a final embedding vector. This process is termed multi-head attention. Using multi-head attention allows the transformer to learn different representations of the parts of the sequence which found to increase the performance. 

The KT can also operate without the state token $\mathbf{s}(n)$. However, during development it was determined that the learning performance was improved by incorporating the state into the inputs. This improvement is attributed to the fact that the state token provides the transformer with information regarding the current configuration of the limb, which is exploited in estimating the necessary inputs for the next tip position in the sequence. 


\section{RESULTS}\label{sec: Results}

As is discussed in Sec.\ref{ssec: dataset}, the KT is trained using the episodes of the dataset generated with the forward kinematics analytic model. A total of 80\% of this analytically generated dataset was for training the model and the remaining 20\% was used as the test dataset.The transformer architecture operates on fixed sequence lengths. Therefore, a sequence length of $N=25$ steps is chosen for training the KT. The embedding size is set to 128, and a 12-layer decoder with an 8-head self-attention layer is utilized. 

The desired tendon forces $\mathbf{T}_d(n)$ are masked with zeros during training and the input token at step $n$ is defined as $\bm{\tau}(n) = [\mathbf{s}(n), \mathbf{r}_d(n), \mathbf{0}]$. The loss function given in (7) is the mean square error between the predicted tendon forces $\mathbf{T}_d(n)$ and the analytically computed actual tendon forces $\mathbf{T}_a(n)$ obtained from the training dataset.

$$
\mathcal{L} = \frac{1}{N} \sum_{n=0}^{N-1} [\mathbf{T}_d(n) - \mathbf{T}_a(n)]^2
\eqno{(7)}
$$

\begin{figure}[htbp]
\centerline{\includegraphics[width=0.9\linewidth]{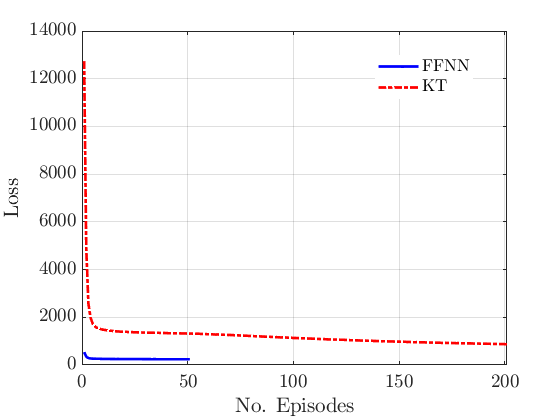}}
\caption{The training loss for the FFNN and the KT.}
\label{fig: trainLoss}
\end{figure}

A conventional FFNN was also trained on the dataset to serve as a baseline inverse kinematics model. The FFNN has two hidden layers of 256 neurons. The input layer is the desired tip position and the output layer is defined as the predicted tendon forces. This FFNN utilized the same loss function given in (7). In contrast to the KT which is trained on sequential data, the FFNN considers each step in the sequence independently from the rest of the data. Both the FFNN and KT models are trained on a NVIDIA GeForce RTX 2060 GPU. The evolution of the loss function for both models are provided in Fig. \ref{fig: trainLoss}. The KT requires more epochs for training as it has more parameters to train, that is why 200 epochs were chosen to train the model while 50 epochs were enough to train the FFNN model.

\begin{figure}[h]
\centerline{\includegraphics[width=\linewidth]{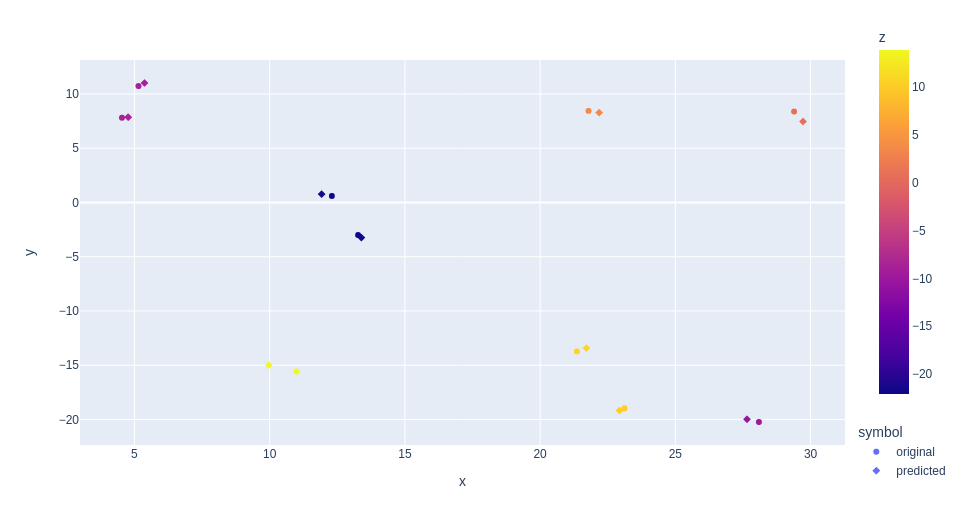}}
\caption{Typical tip positioning results provided by the KT. The squares and circles represent the desired and actual tip positions, respectively. Depth ($z$-axis position) information is color coded.}
\label{fig: tipPosResults}
\end{figure}

A series of numerical simulations are conducted to test the performance of the KT model in predicting the necessary tendon forces. In these simulations, desired tip positions are input to the network that outputs the required tendon forces. These predicted tendon forces are then fed into the analytic forward kinematics model to compute the corresponding actual tip positions. Typical results obtained for the KT are depicted in Fig. \ref{fig: tipPosResults}, showing good overall performance in solving the inverse kinematics problem.  

Furthermore, a benchmark analysis of the baseline FFNN and KT network is performed. The performance of both networks are evaluated in terms of the mean absolute error and standard deviation in predicting the required tendon forces for a given desired tip position in 3-D space. The results are presented below in Table \ref{Tab: forceComp} which indicate that both methods perform uniformly over the four cables. However, the KT produces more accurate and precise estimates of the tendon forces, resulting in a mean absolute error about 30\% less than the FFNN model. In addition to the tendon forces, the resulting positioning errors obtained from the predicted tendon forces is also evaluated and compared in Table \ref{Tab: posComp}. In terms of positioning errors, the KT outperforms the FFNN again produces a mean absolute positioning error less than 5 mm in 3-D space in contrast to the 10 mm provided by the latter.

\begin{table}[h]
\caption{Performance results in terms of mean absolute error and standard deviation of the tendon force predictions for the FFNN and KT networks. All results are presented in N.}
\label{Tab: forceComp}
\begin{center}
\begin{tabular}{|c|c|c|c|c|}
\hline
Model & $T_1 $ & $T_2 $ & $T_3 $ & $T_4 $ \\ 
\hline
FFNN  & $0.81\pm0.7$ & $0.82\pm0.7$ & $0.82\pm0.7$ & $0.81\pm0.7$ \\
\hline
KT  & $0.55\pm0.2$ & $0.55\pm0.2$ & $0.55\pm0.2$ & $0.55\pm0.2$  \\
\hline
\end{tabular}
\end{center}
\end{table}
 
\begin{table}[h]
\caption{Performance results in terms of mean absolute error and standard deviation for the positioning errors for the FFNN and KT networks. All results are presented in mm.}
\label{Tab: posComp}
\begin{center}
\begin{tabular}{|c|c|c|c|}
\hline
Model & $x$-coord. & $y$-coord. & $z$-coord.  \\ 
\hline
FFNN  & $6.1\pm 5.1$ & $4.8\pm 3.7$ & $4.9\pm 4.1$  \\
\hline
KT  & $3.5\pm 3.1$ & $2.5\pm 2.2$ & $2.8\pm 2.5$  \\
\hline
\end{tabular}
\end{center}
\end{table}

 The run-time performance of both methods are also evaluated by measuring th required computational time over several iterations. Accordingly, the FFNN and the KT completed the computations in $115 \pm 6.14 \mu$s and $5920 \pm 189 \mu$s, respectively. The longer computation time associated with the KT is attributed to the higher complexity of the transformer network. 
 
\section{CONCLUSION}\label{sec: Conc}

A machine learning approach for solving the elusive inverse modeling problem of soft robotic limbs is presented in this paper. The proposed KT network is based on the transformer architecture which has gained much interest recently in other application fields outside of robotics. For a given desired tip position in 3-D space, it is shown that the KT outperforms the baseline FFNN-based machine learning approach and can accurately predict the required actuation tendon tensions. Although the scope of the current work is limited to steady-state (kinematic) scenarios, the proposed model can readily be extended to handle sequential multi-waypoint or time-dependent (dynamic) problems for real-time shape and motion control of soft robotic limbs.  


\addtolength{\textheight}{-12cm}   





\section*{ACKNOWLEDGMENT}
This research was supported by The Scientific and Technological Research Council of Turkey (TUBITAK) project no 216M201.


\end{document}